\documentclass[twoside,11pt]{article}

\usepackage{jmlr2e}

\usepackage{xcolor, soul}

\usepackage{amssymb}
\usepackage{todonotes}

\usepackage[flushleft]{threeparttable}
\usepackage{array} 

\usepackage{enumitem}

\usepackage{amsmath}

\ShortHeadings{Neural Network Ensembles: Theory and Training}{Li, Paffenroth and Berthiaume}
\firstpageno{1}

\begin{document}

\title{Neural Network Ensembles: Theory, Training, and the Importance of Explicit Diversity}


\author{\name Wenjing Li \email wli5@wpi.edu \\
  \addr Department of Mathematical Sciences\\
  Worcester Polytechnic Institute\\
  Worcester, MA 01609, USA
  \AND
  \name Randy C.\ Paffenroth \email rcpaffenroth@wpi.edu \\
  \addr Department of Mathematical Sciences, Computer Science and Data Science\\
  Worcester Polytechnic Institute\\
  Worcester, MA 01609, USA
  \AND
  \name David Berthiaume \email dmberthiaume@wpi.edu \\
  \addr Department of Mathematical Sciences\\
  Worcester Polytechnic Institute\\
  Worcester, MA 01609, USA}


\maketitle

\begin{abstract}
Ensemble learning is a process by which multiple base learners are strategically generated and combined into one composite learner. There are two features that are essential to an ensemble's performance, the individual accuracies of the component learners and the overall diversity in the ensemble. The right balance of learner accuracy and ensemble diversity can improve the performance of machine learning tasks on benchmark and real-world data sets, and recent theoretical and practical work has demonstrated the subtle trade-off between accuracy and diversity in an ensemble.  In this paper, we extend the extant literature by providing a deeper theoretical understanding for assessing and improving the optimality of any given ensemble, including random forests and deep neural network ensembles. We also propose a training algorithm for neural network ensembles and demonstrate that our approach provides improved performance when compared to both state-of-the-art individual learners and ensembles of state-of-the-art learners trained using standard loss functions.  Our key insight is that it is better to explicitly encourage diversity in an ensemble, rather than merely allowing diversity to occur by happenstance, and that rigorous theoretical bounds on the trade-off between diversity and learner accuracy allow one to know when an optimal arrangement has been achieved.

\end{abstract}


\section{Introduction}
Ensemble learning is a process by which multiple base learners are generated and combined into one composite learner \citep{dietterich2000ensemble}.
Bagging \citep{nowlan1991evaluation}, boosting \citep{freund1997decision} (including well-known Gradient Boosting \citep{friedman2002stochastic} and Adaboost \citep{freund1997decision}), and stacking \citep{kuncheva2004combining} are classic examples of ensemble algorithms.
Such techniques have been applied to a variety of machine learning domains including text mining \citep{williams2014predicting}, recommender systems \citep{amatriain2011data}, and many others.  One popular and frequently used class of ensemble algorithms are random forests \citep{breiman2001random} which combine decision trees, as base classifiers, with bagging, to provide diversity among the base classifiers. More modern examples of ensembles include ensemble adversarial training \citep{tramer2017ensemble} (an ensemble technique proposed to defend adversarial attacks), and ensemble deep auto-encoders \citep{shao2018novel} (an ensemble method for fault diagnosis of rolling bearings).

A neural network ensemble  \citep{li2018research} is an ensemble that combines the individual outputs of several neural networks, and has numerous machine learning applications including image classification \citep{giacinto2001design}, face recognition \citep{huang2000pose}, and intrusion detection \citep{amini2014effective}, to name but a few. Recent work by \citet{choi2018combining} combines Long Short-Term Memory (LSTM) network ensembles with adaptive weighting to improve time series forecasting, and \citet{nguyen2018acoustic} apply nearest neighbor filters to convolutional neural network ensembles to perform acoustic scene classifications.

\subsection{Related Work}
Over the years researchers have been investigating ways to build good ensembles that outperform their component learners \citep{liu1999ensemble, dietterich2002ensemble, zhang2012ensemble, sagi2018ensemble}. Two features of ensembles which have been shown to be key contributing factors to their performance are the individual accuracies of component learners and the diversity of the learners in the ensemble \citep{kuncheva2004combining}. In particular, there is no single widely accepted metric for measuring the diversity in a given ensemble, but intuitively one can think of a diverse ensemble as one where the failures of one classifier can be compensated for by the successes of the others \citep{sharkey1997combining}, that is, to have a diverse ensemble, the component learners need to avoid making coincident errors.

There also has been a vast amount of research and development on creating diversity in ensembles \citep{kuncheva2003measures}.  In particular, \cite{melville2005creating} proposed a DECORATE algorithm that creates diversity in ensembles by adding randomly generated artificial training examples to the original training data (the details of this algorithm will be introduced later). We implemented their algorithm DECORATE on several non-image data sets, and the results are saved for making future comparisons with our method.

The theorems and numerical results in this paper are inspired by the recent paper \footnote{From December 2019} \citep{icmla_paper} where the authors of that paper demonstrated that there is a subtle trade-off between the individual accuracies and diversity of an ensemble, and that studying this trade-off can lead to both theoretical and practical improvements to machine learning algorithms.  In this paper we extend both the theory and results in \citet{icmla_paper}
and derive a set of algorithms, armed with rigorous theoretical bounds, that improve on the performance of state-of-the-art deep neural networks for both binary and multi-label classification problems.

In many ways, the key results of this paper are quite easy to state.  We merely observe that many approaches to constructing machine learning ensembles only achieve diversity in their base learners by happenstance.  For example, classic random forests \citep{breiman2001random} achieve learner diversity merely by randomly performing bootstrap sampling of their data and randomly choosing subsets of their predictors for their constituent learners.  Gradient boosting \citep{friedman2002stochastic}, and similar methods \citep{freund1997decision}, improve matters by taking the errors of the current set of learners into account when constructing a new learner. \emph{Our proposed algorithm goes even further by explicitly encouraging diversity between the set of learners in the ensemble as part of the training process}, and we use this approach to train ensembles of neural networks where the individual learners already have state-of-the-art performance.  These ensembles then have classification error rates that are \emph{2-6 times lower} than the error rates of the state-of-the-art learners from which they are constructed.

\subsection{Our Contribution}
Herein we develop a new and deeper understanding of the balance between individual accuracies and ensemble diversity, building on \citet{icmla_paper}, by providing a new proof of one important theorem based upon the Cauchy-Schwarz inequality \citep{steele_2004}.  This new proof provides valuable insights into connections between our work, coding theory \citep{irvine2001data}, and Welch bounds \citep{datta2012geometry}.  In addition, we demonstrate how this theoretical work provides methods to assess and improve the optimality of any given classification ensemble (including random forests and ensembles of deep neural networks) by relating the simple majority vote accuracy of ensembles to the learner-learner correlations (which measure ensemble diversity) and truth-learner correlations (which measure individual learner accuracies). Furthermore, we propose an algorithm to train ensembles of deep neural networks and demonstrate the effectiveness of our approach with a variety of experiments on standard benchmark data sets.  In particular, using our methods one is able to start from state-of-the-art neural networks and produce ensembles that improve their accuracy beyond their current capabilities. Moreover, our training algorithm is quite efficient in that we only need to train the pretrained base neural networks, as part of the ensemble, by a small number of additional epochs to achieve good performance.

In summary, in this paper, we make the following novel contributions:
\begin{itemize}
  \item We provide a more concise and illuminating proof of a theorem from the literature, and extend its applicability from binary classification to multi-label classification.
  \item We develop a new theorem that provides a closed-form formula of calculating the simple majority vote accuracy of ensembles based on the statistical correlations between pairs of learners and correlations between the learners and the ground truth.
  \item This new theoretical work offers a clear understanding of how individual accuracies of component learners and diversity in an ensemble affect the performance of an ensemble, and thus inspires a metric to assess the optimality of given ensembles.
  \item We propose a training algorithm for deep neural network ensembles that explicitly encourages ensemble diversity by the way of maximizing our proposed metric.
  \item Our algorithm is demonstrated to be more effective than related algorithms by \cite{melville2005creating} in creating diversity in neural network ensembles and maintaining the overall accuracy in the meantime.
  \item Our algorithm is shown to be general and efficient in improving state-of-the-art neural networks by experiments on both non-image and image classification problems with binary and multi-label data.
\end{itemize}

\section{Accuracy-Diversity Trade-off in Ensembles from the Perspective of Statistical Correlations}
There is a large extant literature \citep{brown2005managing, brown2004diversity, brown2010good, dai2017considering, asafuddoula2017incremental, hsu2017theoretical} that has studied the relationship between the individual accuracies of component learners and diversity in ensembles. One recent work by \citet{icmla_paper} showed the balance between these two features from the perspective of Pearson statistical correlations \citep{benesty2009pearson}. In particular, the term ``averaged learner-learner correlations" denoted by $r_{LL}^{(ave)}$ was defined as follows as a measure for the overall diversity in an ensemble \citep{icmla_paper}:
\begin{equation*}
  r_{LL}^{(ave)}=\frac{1}{N(N-1)/2}\sum_{i=1}^{N}\sum_{j >  i}^{N}r_{L_i, L_j},
\end{equation*}
where $N$ is the ensemble size (i.e. the number of learners in the ensemble), and $r_{L_i, L_j}$ is the pairwise Pearson correlation coefficient between the predicted labels of learners $L_i$ and $L_j$. Similarly the term ``averaged truth-learner correlations" noted by $r_{TL}^{(ave)}$ was defined as follows as a measure for the overall accuracy of learners in an ensemble \citep{icmla_paper}:
\begin{equation*}
  r_{TL}^{(ave)}=\frac{1}{N}\sum_{i=1}^{N}r_{T, L_i},
\end{equation*}
where $N$ is again the ensemble size, and $r_{T, L_i}$ is the Pearson correlation coefficient between the labels of ground truth $T$ and the predicted labels of learner $L_i$.

\begin{figure}[htbp]
  \centering
  \includegraphics[width=0.25\linewidth]{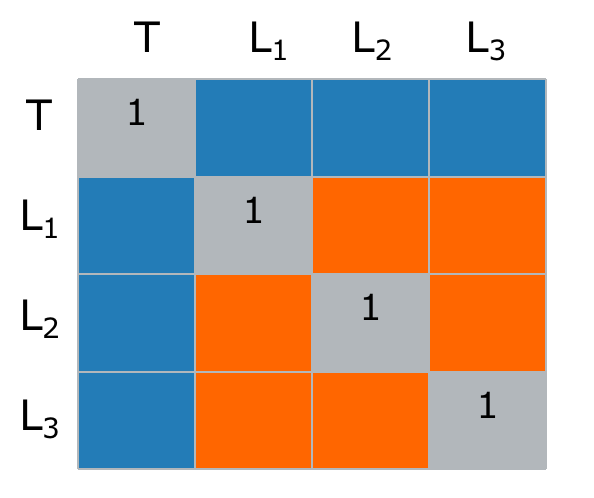}
  \caption{Correlation matrix for the ground truth $T$ and three learners $L_1$, $L_2$ and $L_3$. The average of the blue elements is the averaged truth-learner correlations $r_{TL}^{(ave)}$, and the average of the orange elements is the averaged learner-learner correlations $r_{LL}^{(ave)}$.}
  \label{correlation_matrix}
\end{figure}

To motivate our approach we show in Figure $\ref{correlation_matrix}$ a schematic of the correlation matrix between the ground truth $T$ and three learners $L_1$, $L_2$ and $L_3$.  In this schematic one can see that $r_{TL}^{(ave)}$ and $r_{LL}^{(ave)}$ are calculated by taking the average of the blue and orange elements, respectively.

As discussed earlier, the component learners in a diverse ensemble should not be too similar. Therefore small, or even negative, averaged learner-learner correlations are preferred. Hence a lower value of $r_{LL}^{(ave)}$ indicates a higher level of diversity in the ensemble.  Similariy, since $r_{T, L_i}$ measures the correlation between a learner and the truth (which will be discussed in detail in Section 3), the higher the value of $r_{TL}^{(ave)}$ is, the higher the overall accuracy of the learners in the ensemble is.

\begin{figure}[htbp]
  \centering
  \includegraphics[width=0.53\linewidth]{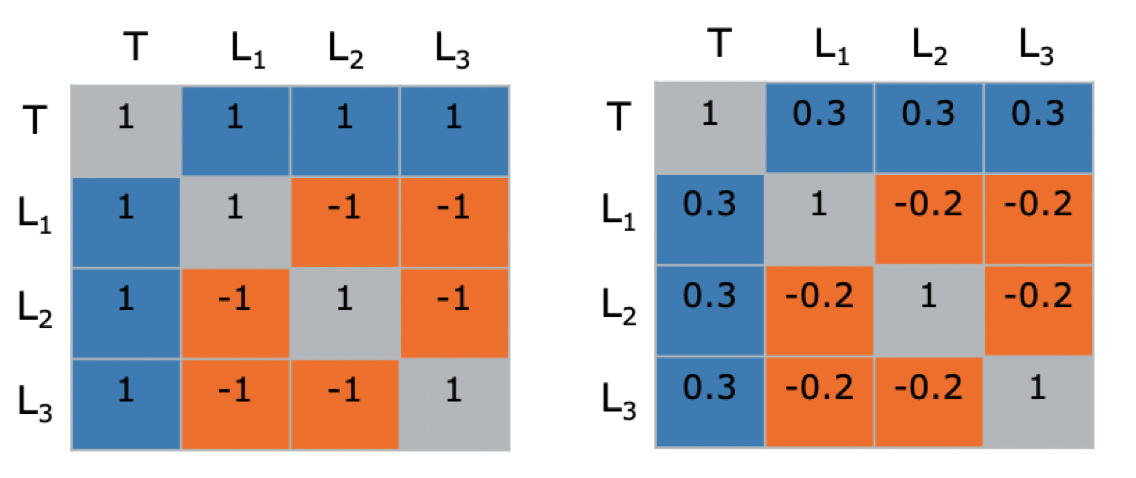}
  \caption{Two examples of a correlation matrix for the ground truth $T$ and three learners $L_1, L_2$ and $L_3$. The left one is not a valid correlation matrix, as it is not non-negative definite, while the right one is valid. These two examples show the fact that not all values between $-1$ and $1$ are possible for the averaged truth-learner correlations and the averaged learner-learner correlations.}
  \label{examples_correlation_matrix}
\end{figure}

Accordingly, one may be tempted to consider the left correlation matrix in Figure $\ref{examples_correlation_matrix}$ as optimal for the three-learner cases, as its $r_{LL}^{(ave)}$ has reached the lowest possible value $-1$ and its $r_{TL}^{(ave)}$ has reached the highest possible value $1$. However, \emph{no such correlation matrix can exist}, as it is not non-negative definite and thus is not a legitimate correlation matrix.  However, the right correlation matrix in Figure $\ref{examples_correlation_matrix}$ is valid with $r_{TL}^{(ave)}=0.3$ and $r_{LL}^{(ave)}=-0.2$ since all of its eigenvalues are greater than or equal to $0$.

Generalizing these ideas by studying the eigenstructure of correlation matrices leads to the two theorems below, originally proven in \citet{icmla_paper}, which provide sharp bounds on what values of $r_{TL}^{(ave)}$ and $r_{LL}^{(ave)}$ are possible for ensembles of different sizes.

\begin{theorem}(Originally from \citet{icmla_paper} where it appears as Theorem 1)
  \label{thm1}
  For an ensemble with $N$ learners we have that
  \begin{equation}
    -\frac{1}{N-1}\leqslant r_{LL}^{(ave)} \leqslant 1.
    \label{theorem1}
  \end{equation}
\end{theorem}

\newpage

\begin{theorem}\label{thm2} (Originally from \citet{icmla_paper} where it appears as Theorem 2)
  For an ensemble with $N$ learners we have that
  \begin{equation}
    -\sqrt{\frac{(N-1)\cdot r_{LL}^{(ave)}+1}{N}}\leqslant r_{TL}^{(ave)} \leqslant \sqrt{\frac{(N-1)\cdot r_{LL}^{(ave)}+1}{N}}.
    \label{theorem2}
  \end{equation}
\end{theorem}

Theorem $\ref{thm1}$ shows that there is a limitation to the extent of possible negative learner-learner correlations in an ensemble. In particular, as the ensemble size $N$ increases, the lower bound $-\frac{1}{N-1}$ in $\eqref{theorem1}$ approaches $0$ from the negative side, meaning that larger ensembles cannot have large negative values of $r_{LL}^{(ave)}$ as compared to smaller ensembles.   Along these same lines, Theorem $\ref{thm2}$ provides the tight bounds for $r_{TL}^{(ave)}$ as functions of $r_{LL}^{(ave)}$.

The original proof in \citet{icmla_paper} to Theorem $\ref{thm2}$ followed the idea of Sylvester's criterion \citep{gilbert1991positive}.   Herein we propose a new and shorter proof (in Appendix A.) by applying the Cauchy-Schwarz inequality.  This new, more geometric proof, provides connections between these results and classic ideas in coding theory \citep{irvine2001data} such as Welch bounds \citep{datta2012geometry} and equiangular tight frames \citep{sustik2007existence}.

\begin{figure}[htbp]
  \centering
  \includegraphics[width=0.6\linewidth]{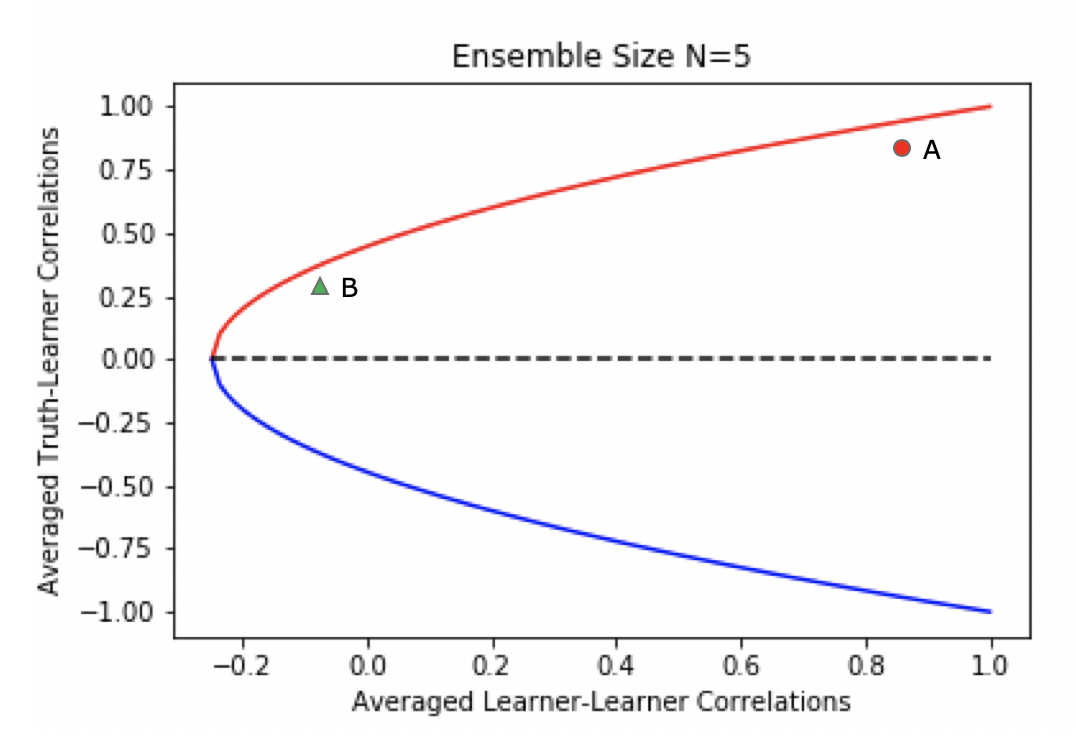}
  \caption{A theoretical plot (averaged truth-learner correlations versus averaged learner-learner correlations) for ensembles of size 5. Ensembles of different sizes have the similar pattern. Any ensemble will fall within the region bounded by the red and blue curve, and an acceptable ensemble should fall at least above the black dotted line.  Ensembles that fall near and on the red curve are optimal, and points A and B are two examples of ensembles that are close to optimal.}
  \label{theorem}
\end{figure}

Perhaps more importantly for the work discussed here, one can combine the two theorems to visualize the relationship between $r_{LL}^{(ave)}$ and $r_{TL}^{(ave)}$.  Figure $\ref{theorem}$ shows an example of this relationship for the cases of an ensemble of size $5$, and ensembles of different sizes will have the similar pattern. In particular, any ensemble will fall within the region bounded by the theoretical upper bound curve (colored in red) and the lower bound curve (colored in blue). An ensemble that is expected to show acceptable performance should at least fall above the black dotted line, where the averaged truth-learner correlations is $0$ (meaning that the component learners are making random guesses on the labels of data instances), as the Condorcet's jury theorem \citep{boland1989majority} states that the individual probabilities being correct must be greater than $1/2$ to ensure the success of the group decision under the rule of simple majority vote.

The red curve in Figure $\ref{theorem}$ can be thought of an optimal boundary, similar to a Pareto boundary \citep{sen1993markets}.  In particular, any ensemble that is strictly below the red line can have either its individual accuracy or diversity improved at no cost to the other quantity.
For example, the two points on Figure $\ref{theorem}$ are examples of ensembles that are close to optimal, where the component learners in the ensemble at point A are very accurate but not diverse, while the learner in the ensemble at point B are very diverse but not that accurate.

\section{Theory for Assessing and Improving the Performance of Ensembles}
In Section 2, we discussed the balance between individual accuracies and diversity of learners in ensembles. In this section, we will demonstrate how exactly the two features affect the performance of ensembles for classification problems. In particular, the simple majority vote accuracy of ensembles will be used as the metric for assessing their performance, and we aim to develop a closed-form formula of calculating simple majority vote accuracy based on the averaged truth-learner correlations $r_{TL}^{(ave)}$ and the averaged learner-learner correlations $r_{LL}^{(ave)}$.

For now, we only derive the theory for simple majority vote accuracy for homogenous ensembles (which will be defined in Section 3.2). However, the lessons learned from the theory for homogeneous ensembles inspire us on developing approaches that are still effective for general ensembles, though without the same theoretical guarantees. In Section 4, we will demonstrate how such ideas lead to our training algorithm for general neural network ensembles.

\subsection{Relationship between the Accuracy of a Learner and its Correlation with the Truth}
Before investigating the theory for assessing the performance of ensembles, it is worthwhile to derive the mathematical formulation of the individual accuracy of a learner in terms of its correlation with the truth, as such formulation is critical to the formulation of simple majority vote accuracy of ensembles (which we will detail in Section 3.2).

For simplicity, consider a data set with two classes (the labels are indicated by $0$'s and $1$'s). The mathematical relationship for binary classification problems between the accuracy of a learner $L$ and its correlation with the true class labels $T$ can be characterized by the following theorem (Proof in Appendix B.). Note that extending the theorem to multi-label classification is one important direction for future work.
\begin{theorem}\label{thm3}
  \begin{equation}
    r_{T, L}=\frac{\beta p - \alpha (2\beta p - p+1-\alpha)}{\sqrt{\alpha(1-\alpha)(2\beta p - p+1-\alpha)(-2\beta p +p+\alpha))}}\label{theorem3},
  \end{equation}
  where
  \begin{itemize}
    \item the learner $L$ and the truth $T$ are both vectors of $0$'s and $1$'s,
    \item $r_{T, L}$ is the Pearson correlation coefficient between $L$ and $T$,
    \item $p$ is the accuracy of learner $L$ (i.e. the proportion of data points correctly classified by learner $L$),
    \item $\alpha$ is the proportion of $1$'s in the truth class labels $T$ (note that $0<\alpha<1$ \footnote{Note that $\alpha$ cannot be 0 or 1, as correlation is not defined for constants (when true class labels are all $0$'s or all $1$'s).}, and for balanced data sets, $\alpha=\frac{1}{2}$),
    \item $\beta$ is the ratio of accuracy in class $1$ to accuracy in both classes (i.e. $\beta=P(T=1\cap L=1)/(P(T=1\cap L=1)+P(T=0\cap L=0))=P(T=1\cap L=1)/p$, and $P(.)$ represents probability).\end{itemize}
  Proof:  In the Appendix.
\end{theorem}

The relationship between $p$ and $r_{T, L}$ appears to be complicated in \eqref{theorem3}. However one can solve $p$ from \eqref{theorem3} by writing out the Taylor series expansion \citep{dienes1957taylor} of $p$ at $\beta=\frac{1}{2}$ as shown in \eqref{taylorexpansion}:
\begin{align}
  p= & 2\alpha(1-\alpha)(1+r_{T, L})-4(\alpha-2\alpha^2+2\alpha^3 \nonumber                                     \\
     & +2\alpha r_{T, L}-6\alpha^2 r_{T, L}+4\alpha^3 r_{T, L}+\alpha r_{T, L}^2-3\alpha^2 r_{T, L}^2 \nonumber \\
     & +2\alpha^3 r_{T, L}^2)(\beta-\frac{1}{2})+O((\beta-\frac{1}{2})^2),\label{taylorexpansion}
\end{align}
\noindent where $O((\beta-\frac{1}{2})^2)$ represents all the higher-order (including the second-order) terms of $(\beta-\frac{1}{2})$.
\eqref{taylorexpansion} can be further simplified as
\begin{equation}
  \begin{aligned}
    p & =2\alpha(1-\alpha)(1+r_{T, L}) \indent (\text{Under the assumption of } \beta=1/2) \\
      & =2\alpha(1-\alpha)+2\alpha(1-\alpha)r_{T, L}.
  \end{aligned}
  \label{linear}
\end{equation}

Based on \eqref{linear}, we can see that as long as the learner $L$ performs equally well on both classes (i.e., $\beta=1/2$) then the accuracy of the learner, $p$, is a linear function of its correlation with the truth $r_{T, L}$. Notice that the intercept and slope in \eqref{linear} share the same function, $2\alpha(1-\alpha)$. Since $0<\alpha<1$, the slope has $2\alpha(1-\alpha)>0$. Therefore, for any learner that is expected to perform equally well on both classes, the individual accuracy $p$ is positively linearly correlated with its correlation with the truth. Hence, the higher the value of correlation $r_{T, L}$ is, the more accurate the learner is.

For a balanced binary data set (i.e. $\alpha=1/2$), \eqref{linear} can be further reduced to
\begin{equation}
  \label{linear_simpler}
  p=\frac{1}{2}+\frac{1}{2}\cdot r_{T, L}. \indent (\text{Under assumptions of  }\beta=1/2, \alpha=1/2)
\end{equation}

\begin{figure}[h]
  \centerline{\includegraphics[width=0.6\textwidth]{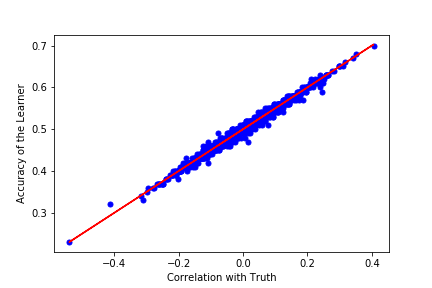}}
  \caption{Accuracy of a Learner versus its Correlation with the Truth. Each of the 2000 blue points is generated by a pair of randomly generated binary classifier and true class labels of size 100 with two classes. All the blue points show a roughly linear trend as indicated by the red straight line.}
  \label{taylor}
\end{figure}

Figure \ref{taylor} visualizes the relationship between accuracy of a learner (the y-axis) and its correlation with the truth (the x-axis) for $2,000$ randomly generated pairs of true class labels $T$ and learner $L$, when no assumptions about $\alpha$ and $\beta$ are made. For each of the $2,000$ pairs, the correlation $r_{T, L}$ and accuracy $p$ are calculated, and marked as a blue dot in Figure \ref{taylor}. All the blue dots show a roughly linear trend (colored in red), where the accuracy is increasing when the correlation increases. There exists some fluctuation of the blue dots around the linear trend, and such fluctuation is due to the higher order terms brought by parameters $\alpha$ and $\beta$ to \eqref{theorem3}.

In conclusion, the fact that a learner’s accuracy is linearly positively correlated with its correlation with the truth is rigorously true if the learner is assumed to perform equally well on predicting both class labels, and is empirically observed even when the assumption is not met. Hence it is reasonable for us to take the truth-learner correlation $r_{T, L}$ as a measure for a learner's individual accuracy. Extending this idea to ensembles, it is also reasonable to take the averaged truth-learner correlations $r_{TL}^{(ave)}$ as a measure for the overall accuracy level of learners in an ensemble.

\subsection{Simple Majority Vote Accuracy of Homogenous Ensembles}

In this section, we will derive the closed-form formula of calculating simple majority vote accuracy of ensembles. In particular, we focus on the concept of ``homogenous ensemble" which is an extension to ensemble learning of the jury design mentioned in \citet{kaniovski2011optimal}, where the authors defined ``homogeneous jury" as a jury in which each vote has an equal probability of being correct, and each pair of votes correlates with the same correlation coefficient. Similarly, we define the ``homogenous ensemble" as an ensemble whose component learners correlate with the true class labels with the same correlation coefficient, and all pairwise learner-learner correlations are equal.

Consider a homogenous jury under the simple majority vote decision rule, a closed-form formula for calculating the probability that the jury being correct (denoted by $M_n(p,c)$) is derived in \citet{kaniovski2011optimal} as follows
\begin{equation}
  \label{eq-reference}
  M_n(p,c)=\sum_{i=\frac{n+1}{2}}^{n}C_n^i p^i (1-p)^{n-i}+0.5c(n-1)(0.5-p) \frac{p^{\frac{n+1}{2}-1}(1-p)^{\frac{n+1}{2}-1}}{B(\frac{n+1}{2},\, \frac{n+1}{2})},
\end{equation}
where $n$ is the number of jurors in the jury, $p$ is the probability of each juror being correct, $c$ is the correlation coefficient of each pair of votes, and $B(.\ , .)$ is Euler's Beta function \citep{chaudhry1997extension}.

Thinking each juror in the homogenous jury as a learner in a homogenous ensemble, inspired by $\eqref{eq-reference}$, and replacing $p$ in $\eqref{eq-reference}$ with $2\alpha(1-\alpha)(1+r_{T, L})$ as derived in $\eqref{linear}$, we are able to derive the formula for calculating the simple majority vote accuracy of ensembles (denoted by $M_{N}(r_{T, L}, r_{L, L})$) as follows
\begin{equation}
  \label{eq-majority-vote}
  \begin{aligned}
    \hspace{0.3in}M_{N}(r_{T, L}, r_{L, L}) &                                                                                                                                                                                                                 \\
                                            & \hspace{-1.3in}=\sum_{i=\frac{N+1}{2}}^{N}C_N^i (2\alpha(1-\alpha)(1+r_{T, L}))^i (1-2\alpha(1-\alpha)(1+r_{T, L}))^{N-i}+0.5\cdot r_{L, L}(N-1)                                                                \\
                                            & \hspace{-1.3in}\cdot(0.5-2\alpha(1-\alpha)(1+r_{T, L}))\cdot \frac{(2\alpha(1-\alpha)(1+r_{T, L}))^{\frac{N-1}{2}}\cdot (1-2\alpha(1-\alpha)(1+r_{T, L}))^{\frac{N-1}{2}}} {B(\frac{N+1}{2},\, \frac{N+1}{2})}, \\
                                            & \hspace{-1.3in} (\text{Under the assumption of }\beta=1/2)
  \end{aligned}
\end{equation}
where $N$ is the ensemble size, $r_{T, L}$ is the truth-learner correlation, $r_{L, L}$ is the Pearson correlation coefficient between each pair of learners, and $\alpha$ is the proportion of $1$’s in the truth class labels $T$ of $0$'s and $1$'s.

In a given homogenous ensemble with binary classifiers that are expected to perform equally well on predicting both classes (i.e. $\beta=1/2$), $\alpha$ can be easily calculated based on the given data set.  In particular, $r_{T, L}$ and $r_{L, L}$ in $\eqref{eq-majority-vote}$ are computable from outputs of the base learners, hence one is able to estimate the simple majority vote accuracy of the ensemble through the formula given by $\eqref{eq-majority-vote}$.  Even when the assumption $\beta=1/2$ is not met, $\eqref{eq-majority-vote}$ still offers a reference for evaluating the simple majority vote accuracy of homogenous ensembles.

For balanced binary data sets (i.e. $\alpha=1/2$), $\eqref{eq-majority-vote}$ can be further simplified as
\begin{equation}
  \begin{aligned}
    M_{N}(r_{T, L}, r_{L, L}) & =\sum_{i=\frac{N+1}{2}}^{N}C_N^i (0.5(1+r_{T, L}))^i (1-0.5(1+r_{T, L}))^{N-i}                                                                         \\
                              & -0.25\cdot \frac{r_{T, L} r_{L, L}(N-1)}{B(\frac{N+1}{2},\, \frac{N+1}{2})}(0.5(1+r_{T, L}))^{\frac{N-1}{2}}\cdot (1-0.5(1+r_{T, L}))^{\frac{N-1}{2}}. \\
                              & (\text{Under the assumptions of }\alpha=1/2,\ \beta=1/2)
  \end{aligned}
  \label{eq-majority-vote-simplified}
\end{equation}

\begin{figure}[h]
  \centerline{\includegraphics[width=0.63\textwidth]{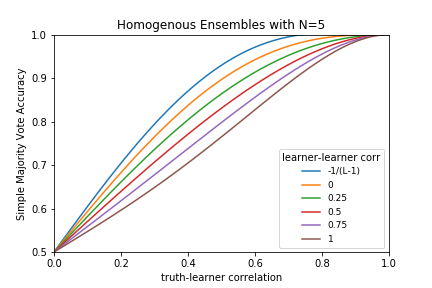}}
  \caption{Simple Majority Vote Accuracy for Homogenous Ensembles of Size 5. Homogenous ensembles of different sizes have the similar pattern. The learner-learner correlation is fixed at six different levels as shown in the legend box, while the truth-learner correlation is varying in $[0, 1]$.}
  \label{majority_vote}
\end{figure}

For fixed values of $r_{L, L}$ in a homogenous ensemble of size $N$, $\eqref{eq-majority-vote-simplified}$ is a function of the single variable $r_{T, L}$, therefore we can visualize the simple majority vote accuracy of the ensembles when $r_{T, L}$ is varying in $[0, 1]$. Figure $\ref{majority_vote}$ is one such example where ensemble size is 5, and the learner-learner correlation $r_{L, L}$ is set at six different levels (as shown in the legend box). Homogenous ensembles of different sizes have the similar pattern.

We can see that from Figure $\ref{majority_vote}$, for fixed level of truth-learner correlation $r_{T, L}$, homogenous ensembles with the lowest possible value of $r_{L, L}$ (indicated by the blue curve) show the highest simple majority accuracy. Moreover, the lower the value of $r_{L, L}$ (that is, the more diverse the ensembles), the higher the majority vote accuracy, which emphasizes the point that one should make the homogenous ensemble more diverse in order to achieve better simple majority vote accuracy, especially in cases that there is not much improvement that might be possible for the individual accuracies of the component learners.  On the other hand, for a fixed level of $r_{L, L}$, the value of simple majority vote accuracy increases as the truth-learner correlation increases, meaning that in order to improve the simple majority vote accuracy of homogenous ensembles, one should try to make the component learners more accurate, in cases where the ensemble cannot be made more diverse.

In summary, one can assess the performance of any given ensemble by evaluating its simple majority vote accuracy. Although the formula $\eqref{eq-majority-vote}$ for calculating simple majority vote accuracy is only valid for homogenous ensembles with certain assumptions, the idea is still useful when we develop approaches for general ensembles in Section 4. More importantly, inspired by the insights obtained from Figure $\ref{majority_vote}$, one may improve the performance of any given ensemble by either making the component learners more accurate, or making them more diverse.

\section{Creating Diversity in Ensembles}
In this section, we will apply the theoretical analysis on ensembles from our earlier sections to practical classification problems. In particular, Theorem \ref{theorem1} and Theorem \ref{theorem2} are general in that they apply to any ensembles whatsoever, and Theorem \ref{theorem3} suggests that $r_{TL}^{(ave)}$ and ensemble accuracy are highly dependent. Accordingly, in this section we study the importance of and ways to create diversity in random forests and deep neural network ensembles from the perspective of statistical correlations ($r_{TL}^{(ave)}$ and $r_{LL}^{(ave)}$).

\cite{melville2005creating} proposed a DECORATE algorithm that creates diversity in ensembles by adding randomly generated artificial training examples to original training data, however, their algorithm is computationally expensive especially for large data sets. We thereafter propose a novel training algorithm following the idea of our theorems introduced earlier, and demonstrate the effectiveness and advantages of our approach in the case of deep neural network ensembles by experimental evaluations on both non-image and image classification problems.  


\subsection{Creating Diversity in Random Forests}
Based on the analysis in earlier sections, we see that diversity plays an important role in the success of ensemble methods. This leads naturally to the question about how to introduce more diversity into ensembles. Herein, we start the analysis and experiments with random forests, as they represent a well-known ensemble method that uses decision trees as component classifiers and is easy to train.

One possible way to introduce diversity into random forests is varying the training parameters of the component decisions trees. Table \ref{random_forest} displays the three types of random forests we generated for the purpose of comparisons. In particular, an original random forest with $5$ trees is trained with default parameters, and is used as our baseline model. The second type is ``feature random forest", where each of the 5 component decision trees is trained with a random subset of features. And the third type is ``depth random forest", where the maximum depth of each tree is specified as a given value. One can imagine generating more types of random forests by varying the other training parameters.
\begin{table}[htbp]
  \renewcommand{\arraystretch}{1.3}
  \caption{Three Types of Random Forests}
  \label{random_forest}
  \centering
  \begin{tabular}{cp{12cm}}
    \hline
    \hline
    Original RF & Original random forest model with $5$ trees using bootstrap sampling of the data for diversity. This is our baseline random forest model for comparison.                                                           \\
    \hline
    Feature RF  & Random forest with $5$ trees, but each tree is trained on a random subset of $m$  features. For the $5$ ``feature random forests" displayed in Figure $\ref{fig_random_forest}$, $m=1, 3, 5, 7, 20$, respectively. \\
    \hline
    Depth RF    & Random forest with $5$ trees with a maximum depth of $d$. For the $5$ ``depth random forests" displayed in Figure $\ref{fig_random_forest}$, $d=3, 5, 7, 9, 11$, respectively.                                     \\
    \hline
    \hline
  \end{tabular}
\end{table}

\begin{figure}[htbp]
  \centerline{\includegraphics[width=0.65\textwidth]{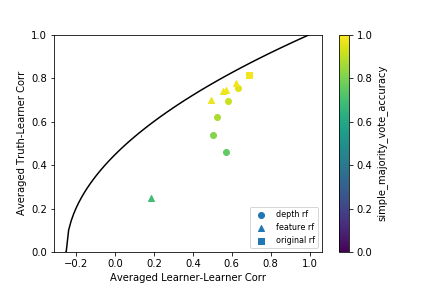}}
  \caption{Three Types of Random Forests Applied on CIFAR-10 train set (check Table $\ref{random_forest}$ for details). There is one original random forest (square shaped), five ``feature rf"  (triangle shaped), and five ``depth rf" (round shaped). The color bar is the simple majority vote accuracy, with the more yellow the points, the higher the accuracy of the forests.}
  \label{fig_random_forest}
\end{figure}

We apply the three types of random forests on the train set of the widely studied CIFAR-10 image data \citep{krizhevsky2009learning} (note:  the training set consists of $50,000$ $32\times32$ color images in $10$ classes including ``car", ``plane", ``ship", etc., with 5,000 images per class) to classify the images, and visualize their calculated metrics ($r_{TL}^{(ave)}$, $r_{LL}^{(ave)}$ and the simple majority vote accuracy) on Figure \ref{fig_random_forest}, where $1$ ``original rf", $5$ ``feature rf" and $5$ ``depth rf" are shown. The color bar is the simple majority vote accuracy, with the more yellow the points, the higher the accuracy of the forests.

We can see from Figure \ref{fig_random_forest}, the ``original rf" has the highest accuracy among all the random forests models, as it is the one closest to the upper bound curve. For ``feature rf" and ``depth rf", the closer they are to the upper bound curve, the higher their simple majority accuracy. ``feature rf" and ``depth rf" are created for the purpose of introducing diversity and outperforming the `'`original rf", but as we can see, the diversity provided by bootstrapping the data is of higher quality than the other types of diversity.

Based on the analysis on random forests, we can see that for ensembles consisting of traditional classification models, even though we can tune their training parameters, it is not guaranteed that such tuning will necessarily improve the performance of the ensembles. However, the theorems we provide here, and our measure of diversity, illuminates the underlying causes of the superior performance of some of the ensembles.

\subsection{Creating diversity in ensembles using artificial data} \label{decorate_section}
To introduce diversity intentionally to ensembles, \cite{melville2005creating} proposed an algorithm DECORATE (Diverse Ensemble Creation by Oppositional Relabeling of Artificial Training Examples), that creates diverse ensembles by adding randomly generated artificial training examples to the original training data. The ensemble is initialized with one existing ``strong" classifier, and then at each iteration a new classifier is trained on the combination of the original training set and a set of artificially generated data (the size of the artificial data is $r$ times the original training size, where $r \in [0,1]$) \citep{melville2005creating}.  For numeric attributes, the new artificial data is randomly generated from a Gaussian distribution of which the mean and standard deviation are computed from the original training set; for nominal attributes, the distribution of the artificial data is obtained by applying Laplace smoothing on the probabilities of occurrences of all the distinct values \citep{melville2005creating}.  The labels of the artificial training examples are then determined in a way such that the probability of the selection is inversely proportional to the predictions of the current ensemble \citep{melville2005creating}. To maintain training accuracy, at each iteration a new classifier will be rejected if adding it to the current ensemble will decrease its accuracy, and iterations can be repeated until an ensemble of desired size or the pre-fixed maximum number of iterations has been reached  \citep{melville2005creating}.

The idea of the DECORATE algorithm is straightforward while attempting to increase diversity in ensembles, by forcing the new classifiers to differ from the current ensemble. However,
a few points with the algorithm may raise doubts on the effectiveness of this method. First of all, their assumptions on the distribution (Gaussian for numeric attributes) of new generated artificial data need further careful discussions. Secondly, at each iteration, a new artificial training set needs to be generated and the probabilities need to be inverted, which is time-consuming and computationally expensive, especially when the original training size is large. Last but not least, whether its success on training data can be generalized to testing data is questionable, as the ensemble diversity is created based on the combination of the artificial and original training data, not purely on the original training data.

In Section \ref{neural network ensemble}, we develop a different way of creating diversity in neural network ensembles, which shows better performance than the DECORATE algorithm.

\subsection{Creating diversity in Neural Network Ensembles} \label{neural network ensemble}

In this section, we will continue working on relating our theoretical results about accuracy and diversity to real-world ensembles. In particular, we propose a training algorithm for deep neural network ensembles, apply the algorithm to a variety of standard benchmark data sets, and demonstrate the effectiveness of the algorithm on both non-image and image classification problems.

\subsubsection{\textbf{Methodology}}
One advantage of neural network ensembles over other traditional ensembles (e.g. random forests) is that we can explicitly control the trade-off between base learner accuracy and diversity in the loss function we use to train the networks through the process of backpropagation \citep{hecht1992theory}. The loss function we proposed, inspired by \citet{icmla_paper}, is one that takes both the accuracy and diversity of neural network ensembles into consideration.
\begin{equation}
  \label{eq-loss}
  Loss=-(r_{TL}^{(ave)}-\lambda\cdot r_{LL}^{(ave)}),
\end{equation}
where $r_{TL}^{(ave)}$ and $r_{LL}^{(ave)}$ is taken as a measure for overall accuracy and diversity of ensembles, respectively.

Given the above loss function, we propose a training algorithm for neural network ensembles that works for both binary and multi-label classifications.  The details of our proposed \textbf{training algorithm} are as follows\\

\newpage
\noindent \framebox[\linewidth][l]{
  \begin{minipage}{\linewidth}
    \begin{itemize}[leftmargin=0pt,label={}]
      \item Input: $X\in R^{n\times q}$, $Y \in R^{n\times m}$, where $n$ is the number of instances, $q$ is the number of\\ features, $m$ ($m>1$) is the number of classes, $Y$ is the true class labels after one-hot \\encoding.

      \item \setlength\itemindent{20pt} For epoch in range(num(epochs)):
      \item optimizer.zero\_grad( )
      \item $O$=[\  [ ] for $j$ in range (ensemble\_size)\ ]
      \item \setlength\itemindent{40pt} for $j$ in range(ensemble\_size):
      \item	 \setlength\itemindent{20pt} $O[j]$=torch.softmax(nets$[j]$($X$), dim=$1$)
      \item $r_{TL}=0, r_{LL}=0$
      \item \setlength\itemindent{40pt} for $k$ in range($m$):
      \item \setlength\itemindent{60pt} for $j$ in range(ensemble\_size):
      \item   \setlength\itemindent{80pt}      $r_{TL}+=Corr(Y[: , k], O[j][:, k])$
      \item  \setlength\itemindent{100pt}  for $i$ in range(ensemble\_size):
      \item  \setlength\itemindent{110pt} if $i<j$:
      \item \setlength\itemindent{20pt} $r_{LL}+=Corr(O[i][: ,k], O[j][:, k] )$
      \item loss=$-(r_{TL}-\lambda\cdot r_{LL})$
      \item loss.backward( )
      \item optimizer.step( )
    \end{itemize}
  \end{minipage}
}\\

In particular, the major steps of the training algorithm for a data set with $m$ classes are:\\
\noindent \textbf{Step 1.} Obtain the one-hot encoded \citep{harris2010digital} true class labels $Y$. \\
\noindent \textbf{Step 2.} Obtain all the corresponding soft outputs $O$ produced from each of the neural networks in the ensemble of size $N$, where each column of $O$ represents the probability for instances to fall into that category, and each row sum of $O$ equals the total probability $1$.\\
\noindent \textbf{Step 3.} Compute all column-wise Pearson correlations between the network outputs $O$ and the true class labels $Y$, for each of the network in the ensemble. Take the overall sum of these column-wise correlations as a measure of truth-learner correlations $r_{TL}$ in the neural network ensemble. \\
\noindent \textbf{Step 4.} Similarly compute all column-wise Pearson correlations between the outputs $O$, for each pair of networks in the ensemble. Take the overall sum of these column-wise correlations as a measure of learner-learner correlations $r_{LL}$ in the neural network ensemble.\\
\noindent \textbf{Step 5.} The neural network ensemble now can be trained with the loss function
\begin{equation}
  \label{new_loss}
  Loss=-(r_{TL}-\lambda\cdot r_{LL}),
\end{equation}
where $r_{TL}$ (calculated by step 3) is a measure for the learner accuracy in the neural network ensemble (with the higher $r_{TL}$ is, the higher the learner accuracy is), and $r_{LL}$ (calculated by step 4) is a measure for the diversity (with the lower $r_{LL}$ is, the more diverse the ensemble is).

Notice that in the loss function $\eqref{new_loss}$, the outer-most minus sign is introduced, as we aim to minimize the training loss by maximizing $r_{TL}$ and in the meantime minimizing $r_{LL}$. Backpropagation can easily be performed with this loss function, as all the parts in $\eqref{new_loss}$ are smooth and differentiable. More importantly, the parameter $\lambda$ ($\lambda>0$) controls the diversity level imposed on the neural network ensemble, with the larger $\lambda$ is, the more diverse the networks in the ensemble will be.

\emph{The ability to explicitly control the level of diversity is a significant advantage of our training algorithm, as opposed to other current published work regarding neural network ensembles that merely allow diversity to occur by happenstance. Another advantage of our training algorithm is that it is generally applicable to neural network ensembles of any size and any network architecture, as all parts in the loss function $\eqref{new_loss}$ are computable with no restrictions on neural network architecture.}

\subsubsection{\textbf{Experiments on Non-Image Data}}  \label{non-image}
In order to show the effectiveness of our proposed training algorithm on non-image data classification problems, perhaps more importantly, to show the advantage of our algorithm over the DECORATE algorithm \citep{melville2005creating} introduced in section \ref{decorate_section},  herein we run experiments on three UCI non-image data sets \citep{Dua:2019} used by \cite{melville2005creating}: Breast Cancer Wisconsin (Original) Data Set, Iris Data Set, and Image Segmentation Data Set. The detailed information of the three data sets is summarized in Table \ref{data_summary}.

We compared the performance of the neural network ensemble trained with our algorithm (with $\lambda=0.9$), the neural network ensemble trained with DECORATE, and the original J48-based DECORATE ensemble used in \cite{melville2005creating} \footnote{The Python code for the experiments in Section \ref{non-image} can be found on \url{https://bitbucket.org/wli5/neural-network-ensembles-theory-training-and-the-importance-of/src/master/}.}. Note that in \cite{melville2005creating} the DECORATE algorithm has not been applied to neural networks, therefore herein we apply it to neural network ensembles, which will be compared to the neural network ensembles trained with our algorithm. To have a fair comparison, the three ensemble methods share the same ensemble size $15$, and for DECORATE methods, the amount of artificial data generated is set to be equal to the size of the original training set. The prediction performance of each of the three ensemble methods are evaluated using $10$-fold cross validation \citep{hastie2005elements}, that is, each data set in Table \ref{data_summary} is randomly split into $10$ equal-size segments of which $9$ segments are used for training and the other one is used for testing. The prediction error rate is then averaged over the $10$ trials.

\begin{table}[htbp]
  \renewcommand{\arraystretch}{1.3}
  \caption{Summary of Non-Image Data Sets}
  \label{data_summary}
  \centering
  \begin{tabular}{cccc}
    \hline
    \hline
    Name     & Instances & Classes & Attributes \\
    \hline
    Breast-w & 699       & 2       & 9          \\
    Iris     & 150       & 3       & 4          \\
    Segment  & 2310      & 7       & 19         \\
    \hline
    \hline
  \end{tabular}
\end{table}

The $10$-fold cross validation error rates of the three ensemble methods on the three data sets are tabulated in Table \ref{DECORATE_comparison}. We can see that out of the three non-image datasets, for both Breast-w and Iris data, the neural network ensemble trained with our algorithm outperforms the DECORATE ensembles, and the error rate on Breast-w even gets reduced by half when using our algorithm compared to the original DECORATE ensemble. For Segment data, our method still shows competitive performance, and gives a lower error rate than the DECORATE neural network ensemble.

\begin{table}[htbp]
  \renewcommand{\arraystretch}{1.3}
  \caption{$10$-fold Cross Validation Error Rate \% (DECORATE Versus Our Algorithm)}
  \label{DECORATE_comparison}
  \centering
  \begin{tabular}{p{9cm}|c|c|c}
    \hline
    \hline
                                                   & Breast-w        & Iris            & Segment         \\
    \hline
    DECORATE in \cite{melville2005creating} & 3.69            & 5.33            & \color{red}1.97 \\
    DECORATE on Neural Network Ensemble            & 2.04            & 10.67           & 3.90            \\
    Our Algorithm on Neural Network Ensemble       & \color{red}1.74 & \color{red}4.00 & 3.33            \\
    \hline
    \hline
  \end{tabular}
\end{table}

\begin{figure}[htbp]
  \centerline{\includegraphics[width=0.5\textwidth]{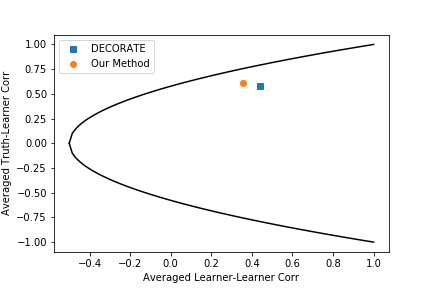}}
  \caption{Training on Breast-w Data. Our neural network ensemble (in orange) shows an overall higher diversity and higher accuracy than DECORATE neural network ensemble (in blue).}
  \label{decorate}
\end{figure}

We can take a closer look at the training process of the neural network ensemble trained with DECORATE and our algorithm, respectively, by visualizing their computed averaged truth-learner correlations and averaged learner-learner correlations on the same graph. Figure \ref{decorate} shows such an example when the two neural network ensembles are trained on the Breast-w data set. Note that following the analysis in previous sections, our ensemble exhibits an overall higher diversity and higher accuracy than the DECORATE ensemble, and since our ensemble is closer to the optimal boundary, it is guaranteed by theory that our ensemble should show a better prediction performance, which is exactly the case as we look at the 10-fold cross validation error rates on the Breast-w data.

In summary, compared to the DECORATE algorithm, our algorithm does a better job on creating diversity in neural network ensembles and keeping the ensembles accurate in the meantime. In addition, the DECORATE algorithm is less efficient and much more expensive computationally, especially when the ensemble size is large as it needs to generate artificial training samples at each iteration.

\subsubsection{\textbf{Experiments on Image Data}}  \label{image}
All our experimental studies included in this section are conducted using the library PyTorch 1.1.0 [22] in Python 3.6.9 [23]. We run the Python scripts on compute nodes with $1$ NVIDIA K20 GPU with $6$ Intel CPU 2.10 GHZ cores and $32$ GB memory each. \footnote{The Python code for the experiments in section \ref{image} can be found on \url{https://bitbucket.org/wli5/neural-network-ensembles-theory-training-and-the-importance-of/src/master/}.}

Herein our focus is on image classification of the benchmark image data sets CIFAR-10 and CIFAR-100 \citep{krizhevsky2009learning}, where we perform training on the train sets and testing on the test sets as provided by the PyTorch library.  However, the theory and training algorithm provided in this paper are also applicable to many other machine learning tasks. To show the general effectiveness of our training algorithm on neural network ensembles, we conduct three major comparison studies (results are displayed in Table \ref{model comparison 1}, \ref{model comparison 2}, \ref{model comparison 3}), and compare the prediction performance of the neural network ensembles trained with our proposed approach and other state-of-the-art individual networks as well as ensembles of state-of-the-art networks.

Table \ref{model comparison 1} shows the comparison study on a binary subset (``plane" and ``car") of the CIFAR-10 image data. In particular, five single pretrained networks (ShuffleNet-v2 \citep{ma2018shufflenet}, GoogLeNet \citep{szegedy2015going}, MobileNet-v2 \citep{sandler2018mobilenetv2}, ResNet-18 \citep{he2016deep} and DenseNet-121 \citep{huang2017densely} included in ``torchvision.models") originally trained on ImageNet \citep{imagenet_cvpr09} are individually re-trained using the standard cross entropy loss function.  The prediction error rate of the five single nets are listed in Table \ref{model comparison 1}. Based on the five single nets, two ensembles are built up: one (displayed as ``Model II" in Table \ref{model comparison 1}) is trained with standard ensemble-version of cross entropy loss (that is, the cross entropy loss of the ensemble is calculated between the averaged outputs from the component nets and the true class labels \citep{icmla_paper}), the other one (displayed as ``Our Model") is trained by our proposed algorithm with various specified $\lambda$ values. We can see that the neural network ensemble trained by our approach outperforms both those single networks and the ensemble trained with standard approach, in fact, our ensemble with $\lambda=0.3$ shows the lowest prediction error rate of $1.25\%$, which is over 2 times lower than the best single net MobileNet-v2, and 6 times lower than the standard ensemble. \emph{This comparison study successfully demonstrates the effectiveness of our training algorithm on binary image classification problems. }

\begin{table}[t]
  \centering
  \begin{threeparttable}
    \renewcommand{\arraystretch}{1.3}
    \caption{Pretrained Networks on CIFAR-10 (Binary Subset: ``Plane" \& ``Car") with Prediction Error Rate (\%)}
    \begin{tabular}{m{2cm}m{6.8cm}c}
      \hline \hline \noalign{\smallskip}
      Model     & Description                                                                                                                                                                                    & Prediction Error Rate (\%) \\
      \hline
      Model I   & Five single re-trained neural networks trained with cross entropy loss until convergence. This is our baseline non-ensemble model for comparison.                                              &
      \begin{tabular}{cc}
        Network       & Error Rate \\
        \hline
        ShuffleNet-v2 & 2.80       \\
        GoogLeNet     & 3.30       \\
        MobileNet-v2  & 2.70       \\
        ResNet-18     & 5.25       \\
        DenseNet-121  & 9.05       \\
      \end{tabular}                                                                                                                                                                                                              \\
      \noalign{\smallskip}\noalign{\smallskip}
      \hline
      Model II  & An ensemble (trained with the ensemble-version of cross entropy loss for $10$ epochs) of the five re-trained nets in Model I .                                                                                              
                & 7.55                                                                                                                                                                                                                        \\
      \hline
      \noalign{\smallskip}\noalign{\smallskip}
      Our Model & An ensemble (trained with our algorithm for $10$ epochs) of the five re-trained nets in Model I. This model demonstrates the practical effectiveness of our approach on binary classification.
                &
      \begin{tabular}{cc}
        $\lambda$ & Error Rate                 \\
        \hline
        0.1       & 1.35                       \\
        0.3       & \textbf{{\color{red}1.25}} \\
        0.5       & 1.70                       \\
        0.7       & 1.90                       \\
        0.9       & 1.80                       \\
      \end{tabular}                                                                                                                                                                                                              \\
      \noalign{\smallskip}\hline\hline
    \end{tabular}
    \begin{tablenotes}
      \small
      \item Note: This table shows the prediction error rates for experiments on a binary subset (``plane" and ``car") of CIFAR-10. The neural network ensemble trained by our approach outperforms both those single networks and the ensemble trained with standard approach, demonstrating the effectiveness of our approach on binary classification.
    \end{tablenotes}
    \label{model comparison 1}
  \end{threeparttable}
\end{table}

Table \ref{model comparison 2} shows the comparison study on the entire CIFAR-10 with all $10$ classes.
Similarly as the study shown in Table \ref{model comparison 1}, three single pretrained networks (ShuffleNet-v2, GoogLeNet and MobileNet-v2) are individually re-trained using the standard cross entropy loss, and two ensembles of size three are built up: one (displayed as ``Model II" in Table \ref{model comparison 2}) is trained with standard ensemble-version of cross entropy loss, the other one (displayed as ``Our Model") is trained by our proposed algorithm with various specified $\lambda$ values. In this case, our ensemble with $\lambda=0.1$ shows the lowest prediction error rate as $5.76\%$, which is much smaller than the best single net GoogLeNet and the standard ensemble. \emph{This comparison study demonstrates the effectiveness of our training algorithm on multi-label image classification problems.}

\begin{table}[t]
  \centering
  \begin{threeparttable}
    \renewcommand{\arraystretch}{1.3}
    \caption{Pretrained Networks on CIFAR-10 with Prediction Error Rate (\%)}
    \begin{tabular}{m{2cm}m{6.8cm}c}
      \hline \hline \noalign{\smallskip}
      Model     & Description                                                                                                                                                                                          & Prediction Error Rate (\%) \\
      \hline
      Model I   & Three single re-trained neural networks trained with cross entropy loss using $40$ epochs. This is our baseline non-ensemble model for comparison.                                                   &
      \begin{tabular}{cc}
        Network       & Error Rate \\
        \hline
        ShuffleNet-v2 & 8.50       \\
        GoogLeNet     & 6.43       \\
        MobileNet-v2  & 7.11       \\
      \end{tabular}                                                                                                                                                                                                                    \\
      \noalign{\smallskip}\noalign{\smallskip}
      \hline
      Model II  & An ensemble (trained with the ensemble-version of cross entropy loss function using $40$ epochs) of the three re-trained nets in Model I.                                                                                         
                & 7.78                                                                                                                                                                                                                              \\
      \hline
      \noalign{\smallskip}\noalign{\smallskip}
      Our Model & An ensemble (trained with our algorithm using 40 epochs) of the three re-trained nets in Model I. This model demonstrates the practical effectiveness of our approach on multi-label classification. &
      \begin{tabular}{cc}
        $\lambda$ & Error Rate               \\
        \hline
        0.1       & \textbf{\color{red}5.76} \\
        0.2       & 5.91                     \\
        0.3       & 6.50                     \\
      \end{tabular}                                                                                                                                                                                                                    \\
      \noalign{\smallskip}\hline\hline
    \end{tabular}
    \begin{tablenotes}
      \small
      \item Note: This table shows the prediction error rates for experiments on the entire CIFAR-10 with all 10 classes. The neural network ensemble trained by our approach outperforms both those single networks and the ensemble trained with standard approach, demonstrating the effectiveness of our approach on multi-label classification.
    \end{tablenotes}
    \label{model comparison 2}
  \end{threeparttable}
\end{table}

A third comparison study is performed to show that with our theory and proposed algorithm, the prediction performance of any given neural network ensembles can be improved even further. Herein, we will take experiments on a recent state-of-art neural network EfficientNet \citep{tan2019efficientnet} as such an example. Table \ref{model comparison 3} shows the prediction error rates of experiments on CIFAR-10 and CIFAR-100 data sets. The baseline model for comparison is a single EfficientNet-b0 net trained with standard cross entropy loss function, while our ensemble model consisting of three copies of the trained EfficientNet-b0 net is trained with our algorithm. We can see that for both data sets, the neural network ensembles trained with our algorithm provide lower prediction error rates than the baseline neural network (which is already very accurate), meaning that \emph{our algorithm is able to make good neural networks even better by explicitly imposing diversity and learner accuracy on the loss function.}
\begin{table}[htbp]
  \centering
  \begin{threeparttable}
    \caption{EfficientNet on CIFAR-10 and CIFAR-100  with Prediction Error Rate (\%)}
    \begin{tabular}{m{2cm}m{7cm}m{2.1cm}m{2.2cm}}
      \hline \hline \noalign{\smallskip}
      Model     & Description                                                                                                & CIFAR-10                                     & CIFAR-100                                   \\
      \hline
      Baseline  & A single EfficientNet-b0 model trained with standard cross entropy loss by $40$ epochs.                    & 5.44\%                                       & 17.28\%                                     \\
      \hline
      Our Model & An ensemble of three copies of re-trained EfficientNet-b0 model trained with our algorithm by $40$ epochs. & \textbf{\color{red}4.65\%} \ ($\lambda=0.1$) & \textbf{\color{red}14.13\%} ($\lambda=0.1$) \\
      \noalign{\smallskip}\hline\hline
    \end{tabular}
    \label{efficientnet}
    \begin{tablenotes}
      \small
      \item Note: This table shows the prediction error rate for experiments on CIFAR-10 and CIFAR-100. For both data sets, the neural network ensembles trained with our algorithm provide lower prediction error rates than the baseline neural network (which is already very accurate), meaning that our algorithm is able to make good neural networks even better.
    \end{tablenotes}
    \label{model comparison 3}
  \end{threeparttable}
\end{table}

Therefore, based on the three comparison studies shown in Table \ref{model comparison 1}, \ref{model comparison 2}, \ref{model comparison 3}, \emph{our training algorithm of neural network ensembles consistently achieves the lowest error rate on a variety of problems and improves the prediction performance of both state-of-the-art individual networks as well as ensembles of state-of-the-art networks.}

\section{CONCLUSIONS}
In this paper, we develop theory that offers a rigorous understanding of how individual accuracies of component learners and diversity affect the performance of an ensemble.  These theoretical considerations inspires methodologies for assessing and improving the optimality of given ensembles.  Just as importantly, we propose a training algorithm for deep neural network ensembles that explicitly encourages ensemble diversity, and this algorithm is shown to be generally effective in improving start-of-the-art neural networks by experiments on standard benchmark data sets.  This training algorithm is also quite efficient in that one only needs to train the ensembles using a small number of epochs to achieve good performance, assuming that the individual learners are already trained outside of the ensemble.

Based upon the results in this paper, there are several interesting directions for future work. First, there appear to be interesting connections between the proof of Theorem \ref{thm2} and domains such as coding theory, and it would be quite useful to understand these connections better.  Second, while our numerical results suggest a strong, and general, connection between $r_{TL}^{(ave)}$ and ensemble accuracy, our Theorem \ref{thm3} only address the case of homogeneous ensembles.  Generalizing this result would likely lead to a deep understanding of the performance of ensembles.  Third, and perhaps most importantly, our numerical experiments beg for application to many more possible machine learning tasks.   There are a wide range of possibilities for taking already state-of-the-art methods and improving them even further by putting them into ensembles.


\newpage

\appendix
\section*{Appendix A.}



\noindent
{\bf Proof to Theorem \ref{thm2}}:\\
Consider the label outputs of the $N$ learners in an ensemble, standardize the outputs so that each output can be considered as a random variable $L_i$ with unit variance, i.e. $Var(L_i)=1, i=1, 2, \cdots, N$. Similarly standardize the labels of the ground truth $T$ such that $Var(T)=1$.

Define the sum of the outputs of the $N$ learners as: $S=\sum_{i=1}^{N} L_i$, then we have
\begin{equation}
  \begin{aligned}
    Cov(S, T) & =Cov(\sum_{i=1}^{N} L_i, T)=\sum_{i=1}^N Cov(L_i, T) \\
              & =\sum_{i=1}^N Corr(L_i, T)=N\cdot r_{TL}^{(ave)},    \\
  \end{aligned}
  \label{eq1}
\end{equation}
where $Cov(.)$ is the covariance, $Corr(.)$ is the correlation.
Note that $Cov(L_i, T)=Corr(L_i, T)$ in $\eqref{eq1}$ because $L_i$'s and $T$ are all unit vectors.

Based on one statistical property of variance, we have
\begin{equation}
  Var(\sum_{i=1}^{N}{L_i})=\sum_{i=1}^{N}Var(L_i)+\sum_{i=1}^{N}\sum_{j\neq i}^{N}Cov(L_i, L_j)\label{eq2}
\end{equation}
Since $Var(L_i)=1$, $i=1, 2, \cdots, N$, the covariance between two random variables equals the correlation, i.e. $Cov(L_i, L_j)=Corr(L_i, L_j)=r_{L_i, L_j}$. We also have $r_{L_i, L_j}=r_{L_j, L_i}$, therefore we can rewrite $\eqref{eq2}$ as
\begin{equation}
  \begin{aligned}
    Var(\sum_{i=1}^{N}{L_i}) & =N+\sum_{i=1}^{N}\sum_{j\neq i}^{N}r_{L_i, L_j}  \\
                             & =N+2\sum_{i=1}^{N}\sum_{j > i}^{N}r_{L_i, L_j}   \\
                             & =N+2\cdot\frac{N(N-1)}{2}r_{LL}^{(ave)}\label{3}
  \end{aligned}
\end{equation}

Finally based on the Cauchy-Schwarz inequality, we get:
\begin{equation}
  \label{eq4}
  \begin{aligned}
    (Cov(S, T))^2 & \leqslant Var (S)\cdot Var(T) \\
                  & =Var(S)\cdot 1                \\
                  & =Var(\sum_{i=1}^{N} L_i)      \\
                  & =N+N(N-1)r_{LL}^{(ave)}
  \end{aligned}
\end{equation}

Combining $\eqref{eq1}$ and $\eqref{eq4}$,
\begin{equation}
  (N\cdot r_{TL}^{(ave)})^2 \leqslant N+N(N-1)\cdot r_{LL}^{(ave)}.
\end{equation}

Therefore, after simplification, we get
\begin{equation}
  -\sqrt{\frac{(N-1)\cdot r_{LL}^{(ave)}+1}{N}}\leqslant r_{TL}^{(ave)} \leqslant \sqrt{\frac{(N-1)\cdot r_{LL}^{(ave)}+1}{N}}
\end{equation}

\section*{Appendix B.}
\noindent
{\bf Proof to Theorem \ref{thm3}}:\\
In this proof, $E(.)$ represents expectation, $Var(.)$ represents variance. Based on basic statistics, the expectation of $T$ is calculated by
\begin{equation}
  E(T)=P(T=1)=\alpha,
\end{equation}
\noindent the expectation of $L$ is calculated by 
\begin{align}
  E(L) & =P(L=1)\nonumber                                    \\
  =    & P(T=1\cap L=1)+P(T=0\cap L=1)\nonumber              \\
  =    & P(T=1\cap L=1) + (P(T=0)-P(T=0\cap L=0))\nonumber   \\
  =    & \beta p+(1-\alpha-(p-\beta p))=2\beta p-p+1-\alpha,
\end{align}
\noindent and the expectation of the product $TL$ is calculated by
\begin{equation}
  E(TL)=P(TL=1)=P(T=1\cap L=1)=\beta p.
\end{equation}
\noindent While the variance of $T$ is given by 
\begin{align}
  Var(T) & =E(T^2)-E(T)^2=P(T^2=1)-P(T=1)^2\nonumber \\
         & =P(T=1)-P(T=1)^2\nonumber                 \\
         & =P(T=1)(1-P(T=1))=\alpha(1-\alpha),
\end{align}
\noindent and the variance of $L$ is given by 
\begin{align}
  Var(L) & =E(L^2)-E(L)^2=P(L^2=1)-P(L=1)^2\nonumber    \\
         & =P(L=1)-P(L=1)^2\nonumber                    \\
         & =P(L=1)(1-P(L=1))\nonumber                   \\
         & =(2\beta p-p+1-\alpha)(-2\beta p +p+\alpha).
\end{align}
\noindent Therefore, based on the definition of Pearson correlation coefficient,\\
\begin{equation}
  \begin{aligned}
    r_{T, L} & =\frac{E(TL)-E(T)E(L)}{\sqrt{Var(T)Var(L)}}                                                                             \\
             & =\frac{\beta p - \alpha (2\beta p - p+1-\alpha)}{\sqrt{\alpha(1-\alpha)(2\beta p - p+1-\alpha)(-2\beta p +p+\alpha))}}.
  \end{aligned}
\end{equation}

\vskip 0.2in
\bibliography{reference}

\end{document}